\pdfoutput=1
\documentclass[11pt]{article}

\usepackage[preprint]{acl}
\usepackage{multirow}
\usepackage{dsfont}
\usepackage{wasysym}
\usepackage{hyperref}
\usepackage{times}
\usepackage{latexsym}
\usepackage{booktabs}
\usepackage{tikz}
\usepackage{graphicx}
\usepackage{tabularx}
\usepackage{graphics}

\usepackage{booktabs}

\usepackage{multirow}

\usetikzlibrary{positioning}
\usepackage{fontawesome5} % For user and bot icons
\usepackage{float}
\usepackage{ragged2e} % For 

\usepackage{adjustbox}
\usetikzlibrary{shapes.callouts, positioning}
\usepackage{xcolor}

\usepackage{amsmath}
\usepackage{amsfonts}
\usepackage{graphicx}
\usepackage{hyperref}
\usepackage[T1]{fontenc}
\usepackage[utf8]{inputenc}

\usepackage{microtype}

\usepackage{inconsolata}

\title{LTDA-Drive: LLMs-guided Generative Models based Long-tail Data Augmentation for Autonomous Driving}
\author{
\normalfont Mahmut Yurt$^{1,2}$ \quad  
Xin Ye$^{1}$\thanks{Corresponding author.} \quad  
Yunsheng Ma$^{1}$ \quad  
Jingru Luo$^{1}$ \quad  
Abhirup Mallik$^{1}$ \\
John Pauly$^{2}$ \quad  
Burhaneddin Yaman$^{1}$ \quad  
Liu Ren$^{1}$ \\
$^{1}$Bosch Research North America \& Bosch Center for Artificial Intelligence (BCAI) \\
$^{2}$Stanford University \\
\texttt{\{myurt,pauly\}@stanford.edu} \quad  
\texttt{\{xin.ye3,jingru.luo,abhirup.mallik,liu.ren\}@us.bosch.com}
}
\begin{document}
\maketitle

\begin{abstract}
3D perception plays an essential role for improving the safety and performance of autonomous driving. Yet, existing models trained on real-world datasets, which naturally exhibit long-tail distributions, tend to underperform on rare and safety-critical, vulnerable classes, such as pedestrians and cyclists. Existing studies on reweighting and resampling techniques struggle with the scarcity and limited diversity within tail classes. To address these limitations, we introduce LTDA-Drive, a novel LLM-guided data augmentation framework designed to synthesize diverse, high-quality long-tail samples. LTDA-Drive replaces head-class objects in driving scenes with tail-class objects through a three-stage process: (1) text-guided diffusion models remove head-class objects, (2) generative models insert instances of the tail classes, and (3) an LLM agent filters out low-quality synthesized images. Experiments conducted on the KITTI dataset show that LTDA-Drive significantly improves tail-class detection, achieving 34.75\% improvement for rare classes over counterpart methods. These results further highlight the effectiveness of LTDA-Drive in tackling long-tail challenges by generating high-quality and diverse data.

\end{abstract} 
\section{Introduction}
\label{sec:intro}
3D perception systems play a critical role in ensuring the safety of autonomous driving, as they provide essential spatial understanding of the environment for downstream prediction and planning tasks~\cite {li_bevformer_2022,jiang_vad_2023,jin2024tod3cap,ma2024mta,ye_bevdiffuser_2025}. These systems are typically trained on real-world datasets that exhibit a long-tailed distribution, where frequent objects (head classes) dominate the scenes, leaving rare objects (tail classes) underrepresented \cite{lvis,liu2019large,Tan_2020_CVPR}. This imbalance poses a significant safety risk, as models trained on such data tend to be biased toward the head classes and underperform on tail classes, particularly on vulnerable groups like cyclists and pedestrians~\cite{peri2023towards,ma2023long}. Addressing this distributional bias is a critical challenge for improving the reliability and safety of autonomous driving technologies.

The class imbalance challenge has traditionally been addressed through reweighting and resampling methods ~\cite{lvis,irfs,Tan_2020_CVPR,Wang_2022_CVPR}. Resampling methods augment long-tail classes by upsampling their instances, whereas reweighting approaches adaptively modify the loss function to emphasize long-tail classes. Although these methods help mitigate the data imbalance problem, they often struggle with the scarcity and lack of diversity within tail classes~\cite{zhao2024ltgc}.

Recently, foundation models such as vision language models (VLMs) and diffusion models have gained significant interest due to their impressive  generation capabilities~\cite{openai_gpt-4_2023,bai_qwen-vl_2023,rombach2022high,podell2023sdxl}. These models have been investigated in a variety of autonomous driving tasks, ranging from perception to planning \cite{mao_gpt-driver_2023,tian_tokenize_2024,hwang2024emma,jin2024tod3cap, jiang2023motiondiffuser,ma2025position}. More recently, they have been used to generate challenging scenarios with scarce weather and lighting conditions, such as rain or nighttime driving \cite{lin2025drivegen}. While these investigations underscore the effectiveness of foundation models for enhancing autonomous driving, their potential to address the long-tail class distribution, where rare and vulnerable classes present  a significant bottleneck, has not been thoroughly explored.

To address the challenges posed by long-tail class distribution in 3D object detection for autonomous driving, we introduce LTDA-Drive, a novel long-tail data augmentation framework, powered by diffusion models and guided by large language models (LLMs). LTDA-Drive augments tail-class objects by generating new and realistic driving scenes from original training images, where head-class objects are replaced in-place with synthesized tail-class objects using diffusion models and evaluated with guidance from LLM agents.

LTDA-Drive performs long-tail data augmentation through three stages. In the first step, removable head-class objects are identified and removed from the training image using a text-guided latent diffusion model. This process removes only the selected head-class objects, preserving the rest of the image to maintain data fidelity. In the subsequent step, diffusion models generate tail-class object instances that align with the regions cleared in the first step. As the final step, leveraging the power of foundation models, an LLM agent filters out low-quality synthesized images, retaining only those that meet the criteria for visual fidelity, semantic consistency, and precise spatial alignment. This comprehensive filtering process also ensures the accurate integration of tail-class objects within the newly proposed 3D bounding boxes. We extensively evaluate our study on the challenging KITTI dataset \cite{geiger2013vision}. The experimental results show that the detectors trained with our augmented data achieve improved detection performance by an average increase of $34.75\%$ for the rare and vulnerable cyclist class, which confirms the LTDA-Drive's remarkable capability in generating high-quality, diverse long-tail data.

We summarize our main contributions as follows:

\begin{itemize}
    \item We introduce LTDA-Drive, a novel LLMs-guided generative model based data augmentation framework that replaces head-class objects with tail-class objects to address long-tail distribution challenges in 3D object detection.
    \item LTDA-Drive introduces three novel modules, namely head-class object removal, tail-class object insertion, and LLM-guided candidate filtering,  to ensure diverse and high-quality long-tail data generation.

    \item Extensive experiments on the challenging KITTI dataset show that detectors trained with LTDA-Drive augmented data achieve significantly improved detection performance of rare classes, demonstrating its ability to generate high-quality diverse long-tail data for downstream tasks.
\end{itemize}

\section{Related Work}
\label{sec:related_work}
\subsection{Long-tailed Detection}
Long-tail distribution remains a key challenge in the realm of computer vision, particularly for object detection tasks. Several works have investigated resampling and reweighting strategies to address the long-tailed detection problem. In resampling, repeat factor sampling techniques have been proposed for increasing the sampling rate of images that contain rare class objects \cite{lvis,irfs}. Another line of work for resampling have focused on mixing multiple object-centric images \cite{zhang2021mosaicos} or copy-pasting long-tail objects on images with the frequent objects \cite{ghiasi2021simple}. In reweighting domain, numerous works with balanced loss strategies have been proposed. Some of the well-known reweighting loss methods for detection includes equalization losses \cite{Tan_2020_CVPR}, category-aware angular margin loss \cite{hyun2022long}, and  effective-class margin loss \cite{Wang_2022_CVPR}. Unlike these works, our work focuses on the scarcity and diversity issue for long-tailed classes by designing an LLM-guided generative models based data augmentation framework.

\subsection{Foundation Models for Autonomous Driving Tasks}

Foundation models such as VLMs and diffusion models    which are pretrained on large scale training data  have gained significant interest for autonomous driving tasks as these models show strong reasoning, generation and generalization capabilities. Several recent works have proposed using vision language models for performing end-to-end planning \cite{hwang2024emma}. Similarly, diffusion models are leveraged for motion prediction~\cite{jiang2023motiondiffuser} and bird's-eye-view (BEV) denoising tasks \cite{ye2025bevdiffuser}. These models have also been extensively used for perception tasks in autonomous driving~\cite{jin2024tod3cap,ma2024mta,lin2025drivegen}. In particular, DriveGen method focuses on long-tailed detection problem by leveraging diffusion model based data augmentation \cite{lin2025drivegen}. Specifically, it focuses on long-tail distribution from weather and time perspective by changing the background of images from sunny/daylight conditions to rainy/night ones. In contrast to DriveGen, our LTDA-Drive focuses on long-tailed detection from class-specific perspective, thus augments rare\&vulnerable classes.

\section{Methodology}
\label{sec:methodology}

\subsection{Overview} 
\begin{figure*}[htb]
    \centering
    \includegraphics[width=0.99\textwidth]{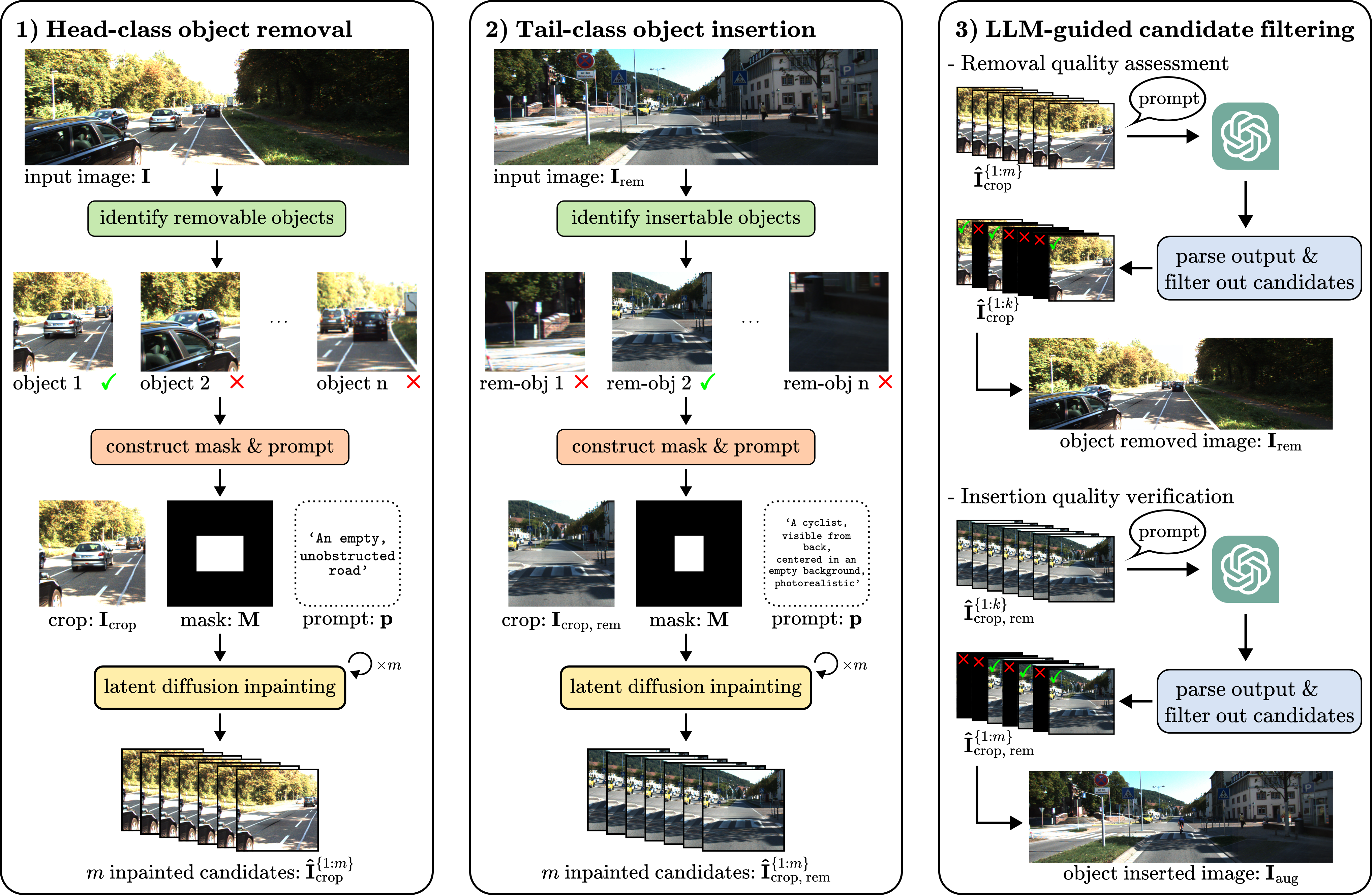}
    \caption{Overview: LTDA-Drive augments data for long-tail classes by replacing frequent objects in driving scenes with rare objects through three key modules. 1) Head-class object removal: identifies removable objects from the original scenes and performs targeted latent diffusion inpainting based on image, mask, and text guidance to generate object-removed candidates. 2) Tail-class object insertion: receives object-removed scenes as input and identifies suitable locations for inserting tail-class objects, followed by localized inpainting to generate candidates. 3) LLM-guided candidate filtering: leverages an LLM agent to assess the quality, plausibility, and geometric alignment of the generated candidates,  to preserve only high-quality generations.}
    
    \label{fig:main_fig}
\end{figure*}

We present \textbf{LTDA-Drive}, a \textbf{L}ong-\textbf{T}ail \textbf{D}ata \textbf{A}ugmentation framework for autonomous driving, designed to address class imbalance and enhance generalization to underrepresented object categories in long-tailed datasets. As illustrated in Fig.~\ref{fig:main_fig}, LTDA-Drive systematically replaces head-class objects with synthetically generated tail-class instances through three key modules: \textbf{1) head-class object removal}, which employs targeted inpainting to erase selected objects; \textbf{2) tail-class object insertion}, which places tail-class objects into semantically coherent empty regions; and \textbf{3) LLM-guided candidate filtering}, which ensures visual realism, contextual alignment, and accurate 3D bounding box placement in the generated images by filtering out low-quality ones. 

\subsection{Head-class Object Removal}
\label{hcor}
To enable the removal of head-class objects, we leverage a localized inpainting method based on a text-conditioned diffusion model built upon SDXL \cite{podell2023sdxl}. Given an input image \( \mathbf{I} \in \mathbb{R}^{H \times W \times 3} \), We select a head-class object for removal based on predefined criteria, ensuring it does not overlap with other objects and meets a minimum size requirement. Subsequently, we extract a square crop centered on the selected object \( \mathbf{I}_{\text{crop}} \in \mathbb{R}^{L \times L \times 3} \) from the image $\mathbf{I}$, where the side length \( L \) is scaled by a factor \( s > 1 \) relative to the object's projected 2D bounding box.

A binary mask \( \mathbf{M} \in \{0,1\}^{L \times L} \) is created over \( \mathbf{I}_{\text{crop}} \) to localize the object, where \( \mathbf{M} = 1 \) within the object's bounding box and \( \mathbf{M} = 0 \) elsewhere. To guide the inpainting process semantically, we provide a text prompt \( \mathbf{p} \in \mathcal{T} \) (e.g., \textcolor{teal}{\texttt{``An empty, unobstructed road''}}) which conditions the generation toward a coherent background.

The masked crop \( \mathbf{I}_{\text{masked}} = \mathbf{I}_{\text{crop}} \odot (1 - \mathbf{M}) \) is first encoded using an encoder \( \mathcal{E} \), yielding a latent representation \( \mathbf{z}_\text{init} = \mathcal{E}(\mathbf{I}_{\text{masked}}) \). This latent \( \mathbf{z}_\text{init} \) is concatenated with the downsampled mask \( \mathbf{M} \) and passed to the denoiser as additional conditioning. During the reverse diffusion process, the latent state \( \mathbf{z}_t \) is iteratively refined at each timestep \( t \) via a text-guided denoising process conditioned on \( \mathbf{p} \). To improve semantic alignment, classifier-free guidance is applied by combining conditional and unconditional noise estimates \cite{ho2022classifier}:
\begin{align}
\hat{\epsilon}_\theta^{\text{cfg}}(\mathbf{z}_t, t, \mathbf{p}, \mathbf{M}) =\; & (1 + w) \cdot \hat{\epsilon}_\theta(\mathbf{z}_t, t, \mathbf{p}, \mathbf{M}) \notag \\
& - w \cdot \hat{\epsilon}_\theta(\mathbf{z}_t, t, \emptyset, \mathbf{M}),
\label{cfg}
\end{align}
where \( w \) is the guidance scale, and \( \hat{\epsilon}_\theta(\mathbf{z}_t, t, \emptyset, \mathbf{M}) \) denotes the unconditional noise prediction.

Once diffusion reaches \( t = 0 \), the final latent \( \mathbf{z}_0 \) is decoded using the decoder \( \mathcal{D} \) to generate the inpainted crop \( \hat{\mathbf{I}}_{\text{crop}} = \mathcal{D}(\mathbf{z}_0) \in \mathbb{R}^{L \times L \times 3} \). This process is repeated \( m \) times to generate multiple candidates \(\hat{\mathbf{I}}_\text{crop}^{\{1:m\}}\), which are then filtered through LLM-guided candidate filtering. Among these, the top \( k \) candidates are selected yielding \(\hat{\mathbf{I}}_\text{crop}^{\{1:k\}}\), and the corresponding crops are seamlessly integrated into the original image \( \mathbf{I} \), replacing the removed object with a contextually appropriate background, resulting in the modified image \( \mathbf{I}_{\text{rem}} \).

\begin{figure}[htb]
\begin{minipage}[t]{\columnwidth}

\begin{tikzpicture}[node distance=0.2cm and 0.1cm]

\node[anchor=west, draw, rounded corners=6pt, fill=gray!10, text width=6.6cm, align=center, font=\small] (q1)
{
\begin{minipage}[t]{6.2cm}
\begin{flushright}\faUser\end{flushright}
\centering
\noindent \hspace{2.2em}{Query} \hspace{4.8em}{Object removal} 

\includegraphics[width=0.47\linewidth]{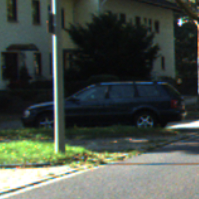}
\hfill 
\includegraphics[width=0.47\linewidth]{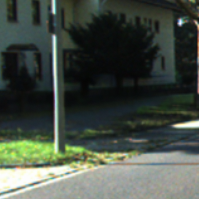}

\vspace{0.5em}
\justifying
\noindent Compare the two images. Was the car removed in the second image without adding or removing any other objects? Respond with a single word: yes or no, followed by a score from 0 to 10 indicating the quality of the object removal (10 = perfect removal, 0 = worst removal). Format your reply as: yes, [score] or no, [score].

\vspace{0.2cm}
\end{minipage}
};

% Assistant response
\node[anchor=center, draw, rounded corners=8pt, fill=green!10, text width=2.2cm, align=left, font=\small, below=of q1, xshift=-3cm] (a1)
{
\begin{minipage}[t]{\linewidth}
\includegraphics[width=1em]{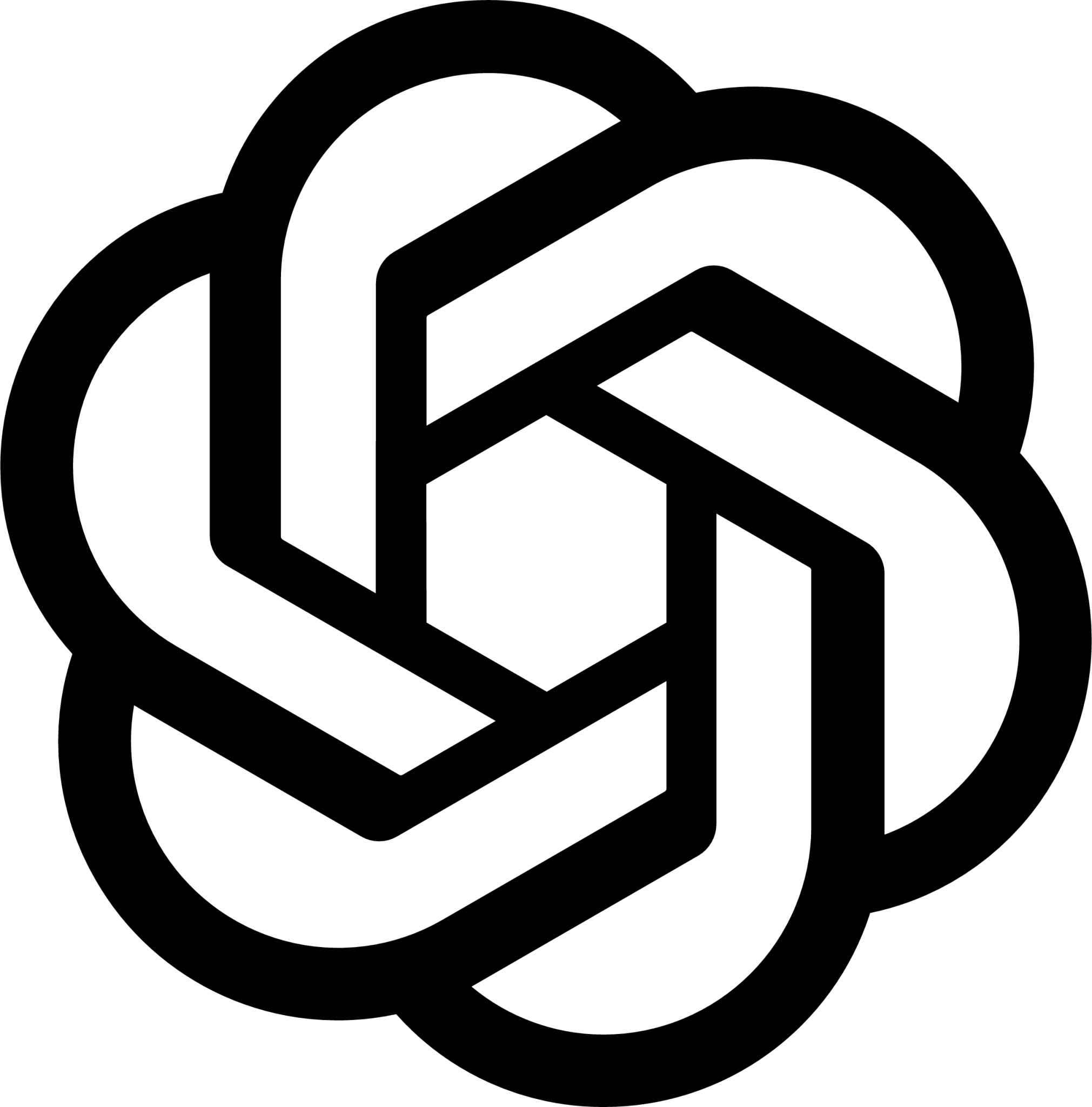}\hspace{0.5em}
\justifying
Yes, 9.
\end{minipage}
};

\end{tikzpicture}
\end{minipage}

\caption{The LLM agent is presented with a pair of images, one containing a head-class object (car) and one after object removal. The agent is asked whether the object was removed cleanly. The response includes a binary yes/no decision and a quality score.}
\label{fig:remove_eval}
\end{figure}

\subsection{Tail-class Object Insertion}
Following the removal of head-class objects, we insert tail-class instances into the cleared regions of the inpainted image \( \mathbf{I}_{\text{rem}} \) to enrich the scene with underrepresented object categories. To preserve spatial and semantic coherence, each tail-class object is inserted at the original spatial center of the removed head-class object.

To assign a realistic bounding box, we sample the tail-class object dimensions from a class-specific statistical distribution. For each dimension \( \text{dim} \in \{h, w, l\} \), denoting height, width, and length, respectively, we draw samples as:

\begin{equation}
\text{dim} \sim \mathcal{N}_{[a_{\text{dim}},\, b_{\text{dim}}]}\left( \mu_{\text{dim}},\, \sigma_{\text{dim}}^2 \right),
\end{equation}
where \( \mu_{\text{dim}} \) and \( \sigma_{\text{dim}} \) are the mean and standard deviation, and \( a_{\text{dim}} \), \( b_{\text{dim}} \) are the empirical minimum and maximum values for the tail-class object dimensions. This ensures that the inserted bounding boxes reflect the natural variation observed in the dataset.
\begin{figure}[htb]
\begin{minipage}[t]{\columnwidth}
\begin{tikzpicture}[node distance=0.2cm and 0.1cm]

% User box - main content container
\node[anchor=west, draw, rounded corners=5pt, fill=gray!10, text width=6.6cm, align=center, font=\small] (q1)
{
\begin{minipage}[t]{6.5cm}
\begin{flushright}\faUser\end{flushright}
\centering
\noindent Positive samples \hspace{1cm} Negative samples\\[0.2em]
\includegraphics[width=.49\linewidth, trim=0 0 130 0, clip]{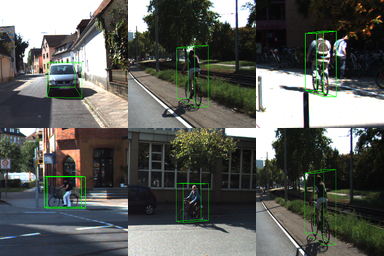}
\includegraphics[width=.49\linewidth, trim=0 0 130 0, clip]{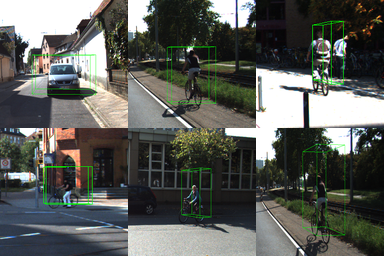}

\vspace{1.5em}
\noindent \hspace{-3.8cm} Query: cyclist image\\[0.2em]

\begin{minipage}[t]{\linewidth}
    \begin{minipage}[t]{0.4\linewidth}
        \includegraphics[width=\linewidth]{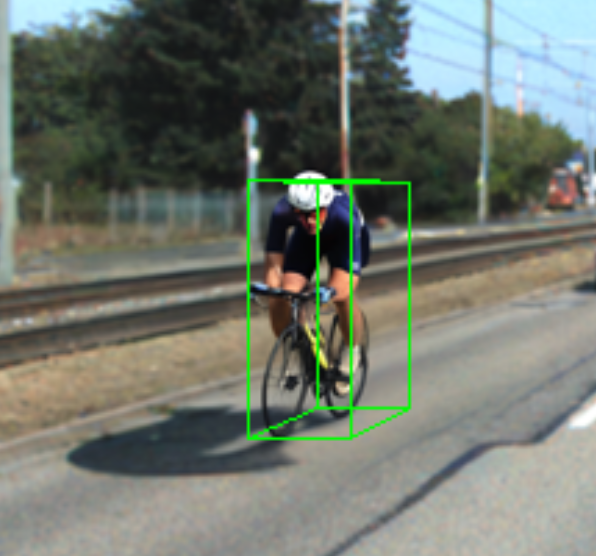}
    \end{minipage}
    \hfill
    \begin{minipage}[t]{0.52\linewidth}
        \raisebox{20ex}[0pt][0pt]{%
        \begin{minipage}[t]{\linewidth}
            \justifying
            \noindent The first image shows examples of well-fitted, accurate 3D bounding boxes. The second image shows loose or poorly aligned boxes. In the third image, does the cyclist fit well within the 3D bounding box? Answer with ‘yes’ or ‘no,’ followed by a brief explanation.
        \end{minipage}
        }
    \end{minipage}
\end{minipage}

\rule{0pt}{2em}

\vspace{0.2cm}
\end{minipage}
};

% Assistant response
\node[anchor=center, draw, rounded corners=8pt, fill=green!10, text width=6cm, align=center, font=\small, below=of q1, xshift=-1cm] (a1)
{
\begin{minipage}[t]{5.8cm}
\includegraphics[width=1em]{sec/figures/in_ctx_lrn/chatgpt_logo.png}\hspace{0.5em}
\justifying
Yes, the cyclist fits well within the 3D bounding box. The box is tightly aligned with the cyclist, with minimal extra space and no significant parts left outside.
\end{minipage}
};

\end{tikzpicture}
\end{minipage}

\caption{The LLM agent is asked to evaluate the 3D bounding box alignment of the inserted tail-class object (cyclist) using in-context learning with positive and negative references. The response includes a binary yes/no decision followed by a brief explanation.}
\label{fig:in_ctx_bbx_eval}
\end{figure}

\begin{figure}[ht]
\begin{minipage}[t]{\columnwidth}

\begin{tikzpicture}[node distance=0.2cm and 0.1cm]

% User box - shift slightly right to align with "Yes"
\node[anchor=west, draw, rounded corners=6pt, fill=gray!10, text width=6.6cm, align=center, font=\small] (q1)
{
\begin{minipage}[t]{6.2cm}
\begin{flushright}\faUser\end{flushright}
\centering
\hspace{0.7em}\noindent{Car} \hspace{7.3em}{Cyclist} 
\includegraphics[width=0.47\linewidth]{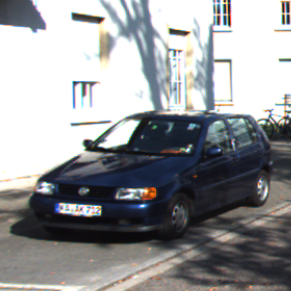}
\hfill
\includegraphics[width=0.47\linewidth]{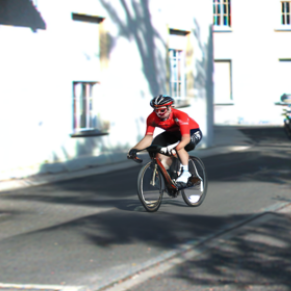}

\vspace{0.5em}
\justifying
\noindent Which side of the car is visible (e.g., front, back, left, right, front-left, back-right, etc.)? Which side of the cyclist is visible (e.g., front, back, left, right, front-left, back-right, etc.)? Are both subjects showing roughly the same side? Begin your answer with 'yes' or 'no,' followed by a brief explanation.

\vspace{0.2cm}
\end{minipage}
};

% Assistant response
\node[anchor=center, draw, rounded corners=8pt, fill=green!10, text width=6.2cm, align=left, font=\small, below=of q1, xshift=-1cm] (a1)
{
\begin{minipage}[t]{\linewidth}
\includegraphics[width=1em]{sec/figures/in_ctx_lrn/chatgpt_logo.png}\hspace{0.5em}
\justifying
Yes, both subjects are showing roughly the same side. the front-left of the car is visible, and the front-left of the cyclist is also visible. Both are facing toward the camera.
\end{minipage}
};
\vspace{-10mm}
\end{tikzpicture}

\end{minipage}

\caption{The LLM agent is provided with a pair of images, one containing the original head-class object (car) and the other one containing the inserted tail-class object (cyclist). The agent is asked to identify if both objects are roughly showing the same side. The response includes a yes/no decision, followed by an explanation.}
\label{fig:orient_eval}
\end{figure}

The 3D bounding box is then projected onto the image plane, yielding a 2D bounding box aligned with the expected position and orientation of the inserted object. A square crop \( \mathbf{I}_{\text{crop, rem}} \subset \mathbf{I}_{\text{rem}} \) is extracted, centered on this 2D box and expanded by a scale factor \( s > 1 \) to include surrounding context. 

Similar to the removal stage, again a binary mask \( \mathbf{M} \in \{0, 1\}^{L \times L} \) is created over \( \mathbf{I}_{\text{crop, rem}} \), where \( \mathbf{M} = 1 \) denotes the region of the projected bounding box designated for tail object synthesis, and \( \mathbf{M} = 0 \) elsewhere. This mask serves as spatial guidance for the generative model.

To semantically steer the synthesis process, we condition the generation on a descriptive text prompt \( \mathbf{p} \in \mathcal{T} \), such as \textcolor{teal}{\texttt{"A \textcolor{blue}{\{tail\_class\_object\}}, visible from \textcolor{blue}{\{orientation\}}, centered in an empty background, photorealistic"}}.

The masked crop \( \mathbf{I}_{\text{masked, rem}} = \mathbf{I}_{\text{crop, rem}} \odot (1 - \mathbf{M}) \) is encoded using an encoder \( \mathcal{E} \), producing a latent representation \( \mathbf{z}_\text{init} = \mathcal{E}(\mathbf{I}_{\text{masked, rem}}) \). As in head-class object removal Section \ref{hcor}, this latent representation is concatenated with the downsampled mask \( \mathbf{M} \) and input to the denoiser network, which iteratively refines the latent state \( \mathbf{z}_t \) through a classifier-free guidance process formulated in Eq. \ref{cfg}.

The final latent of the diffusion process, \( \mathbf{z}_0 \), is decoded through a decoder \( \mathcal{D} \), generating the inpainted patch \( \hat{\mathbf{I}}_\text{crop, rem} = \mathcal{D}(\mathbf{z}_0) \in \mathbb{R}^{L \times L \times 3} \), which contains a photorealistically synthesized tail-class object of interest. Multiple candidates are generated \( \hat{\mathbf{I}}_\text{crop, rem}^{\{1:m\}} \), with low-quality candidates filtered using LLM-guidance, yielding \( \hat{\mathbf{I}}_\text{crop, rem}^{\{1:k\}} \). Finally, the remaining patches are seamlessly blended back into \( \mathbf{I}_{\text{rem}} \), yielding the final augmented image \( \mathbf{I}_{\text{aug}} \).

\subsection{LLM-guided Candidate Filtering}
To ensure visual fidelity and contextual consistency across both removal and insertion outputs, we introduce an LLM agent that serves as a post-generation filtering module. Serving as an automated evaluator, the LLM agent receives as input a set of images and a prompt to assess each candidate prior to final integration.

\subsubsection{Removal Quality Assessment}
Given the localized crop \( \mathbf{I}_\text{crop} \) from the original image and a set of \( m \) inpainted candidates \(\hat{\mathbf{I}}_\text{crop}^{\{1:m\}}\), we evaluate the quality of object removal using an LLM, specifically, ChatGPT-4.1 which serves as the evaluation backbone. Each pair is provided to the LLM with the following prompt: 
\textcolor{teal}{\texttt{"Compare the two images. Was the }}
\texttt{\textcolor{blue}{\{head\_class\_object\}}}
\textcolor{teal}{\texttt{ removed in the second image without adding or removing any other objects? Respond with a single word: yes or no, followed by a score from }}
\texttt{\textcolor{blue}{\{lower\_score\}}} 
\textcolor{teal}{\texttt{ to }} 
\texttt{\textcolor{blue}{\{upper\_score\}}} 
\textcolor{teal}{\texttt{ indicating the quality of the object removal. (\textcolor{blue}{\{upper\_score\}} = perfect removal, \textcolor{blue}{\{lower\_score\}} = worst removal). Format your reply as: yes, [score] or no, [score]."}} The VLM's response, formatted as \textcolor{orange}{\texttt{"yes, \{score\}"}} or \textcolor{orange}{\texttt{"no, \{score\}"}}, yields a semantic quality score for each candidate. A representative example is shown in Fig. \ref{fig:remove_eval}.

From the \( m \) generated outputs, the top \( k \) candidates with the highest scores are retained, ensuring that only artifact-free and contextually consistent backgrounds are preserved for data augmentation.

\subsubsection{Insertion Quality Verification}
Given a set of tail-class insertion candidates \( \mathbf{\hat{I}}_\text{crop, rem}^{\{1:m\}}\), we assess their geometric plausibility and viewpoint consistency using two sequential LLM-based filters.

\begin{table}[ht]
\centering
\caption{Object counts and class distributions across the original and augmented datasets.}
\label{tab:datacount}
\resizebox{\linewidth}{!}{%
\begin{tabular}{lrrrr}
\toprule
\textbf{Split} & \textbf{Car} & \textbf{Cyclist} & \textbf{Pedestrian} & \textbf{Total} \\
\midrule
Original & 14,357 (83.00\%) & 734 (4.24\%) & 2,207 (12.75\%) & 17,298 \\
Augmented & 15,901 (79.64\%) & 1,509 (7.56\%) & 2,555 (12.79\%) & 19,965 \\
\bottomrule
\end{tabular}
}
\vspace{-5mm}
\end{table}

\paragraph{Geometric plausibility via in-context learning.}
To verify the alignment between the inserted object and the generated 3D bounding box, we adopt a few-shot in-context learning approach. The VLM is presented with three exemplars in sequence: (i) a correctly aligned examples, i.e., positive samples, (ii) misaligned examples, i.e., negative samples, and (iii) the current candidate. The LLM agent is then queried as follows:  
\textcolor{teal}{\texttt{``The first image shows accurate 3D bounding boxes, the second shows poor alignment. In the third image, does the }\{\textcolor{blue}{rare\_class\_object}\} fit well within the 3D bounding box?''}  
The LLM agent's response is used to evaluate the geometric plausibility. Among \( m \) generated candidates, only the best candidates are retained to ensure bounding box consistency. A representative example of this process is shown in Fig.~\ref{fig:in_ctx_bbx_eval}.

\begin{table*}[ht]
\centering
{\small
\renewcommand{\arraystretch}{1.25}  % slightly tighter vertical spacing
\begin{tabular}{l|l|ccc|ccc|ccc}
\hline
\textbf{} & \textbf{Method} 
  & \multicolumn{3}{c|}{\textbf{Car}} 
  & \multicolumn{3}{c|}{\textbf{Pedestrian}} 
  & \multicolumn{3}{c}{\textbf{Cyclist}} \\
\cline{3-11}
 &  & Easy & Mod & Hard & Easy & Mod & Hard & Easy & Mod & Hard \\
\hline
\multirow{2}{*}{AP\textsubscript{2D}} 
 & Baseline   & 95.88 & 89.69 & 84.38 & 63.16 & 54.54 & 45.85 & 58.69 & 38.61 & 36.46 \\
 & LTDA-Drive & 96.00 & 90.08 & 82.72 & \textbf{67.58} & \textbf{57.20} & \textbf{50.00} & \textbf{73.91} & \textbf{48.11} & \textbf{45.45} \\
\hline
\multirow{2}{*}{AP\textsubscript{BEV}} 
 & Baseline   & 66.35 & 48.77 & 42.87 & 27.13 & 21.87 & 18.32 & 16.93 & 9.12 & 8.07 \\
 & LTDA-Drive & 64.27 & 48.21 & 42.54 & \textbf{27.55} & \textbf{22.25} & \textbf{18.70} & \textbf{21.51} & \textbf{12.16} & \textbf{11.24} \\
\hline
\multirow{2}{*}{AP\textsubscript{3D}} 
 & Baseline   & 60.35 & 43.69 & 39.09 & 25.63 & 20.43 & 16.97 & 16.21 & 8.99 & 7.92 \\
 & LTDA-Drive & 57.91 & 43.24 & 38.91 & \textbf{27.14} & \textbf{21.86} & \textbf{18.35} & \textbf{21.50} & \textbf{12.15} & \textbf{10.67} \\
\hline
\multirow{2}{*}{AOS} 
 & Baseline   & 95.76 & 89.17 & 83.39 & 56.50 & 48.08 & 40.26 & 43.53 & 28.18 & 26.52 \\
 & LTDA-Drive & 95.36 & 89.19 & 81.48 & \textbf{61.42} & \textbf{50.97} & \textbf{44.15} & \textbf{64.61} & \textbf{40.93} & \textbf{38.72} \\
\hline
\end{tabular}
} % end small
\caption{Comparison of the object detection performance using the original dataset (baseline) and using the augmented dataset (LTDA-Drive). The measurements are presented for car, pedestrian, and cyclists over three difficulty levels, easy, moderate, hard. Bold indicates the best performance on the tail classes.}
\label{tab:quant_res}
\end{table*}

\paragraph{Viewpoint consistency validation.}
To confirm orientational coherence between the removed and inserted objects, we provide the LLM agent with corresponding pairs crops from the original image and the augmented image, each centered at the object's location. The following prompt is issued:  
\textcolor{teal}{\texttt{``Which side of the \textcolor{blue}{\{head\_class\_object\}} is visible (e.g., front, back, left, or right)? Which side of the \textcolor{blue}{\{tail\_class\_object\}} is visible? Are both subjects showing roughly the same side?''} } 
Only the top $k$ candidates demonstrating viewpoint consistency are retained for final augmentation.

Overall, these two VLM-guided filtering stages enforce geometric alignment and viewpoint consistency, ensuring high-quality tail-class insertions within the augmented dataset.

\begin{figure*}[htbp]
    \centering
    \includegraphics[width=0.99\textwidth]{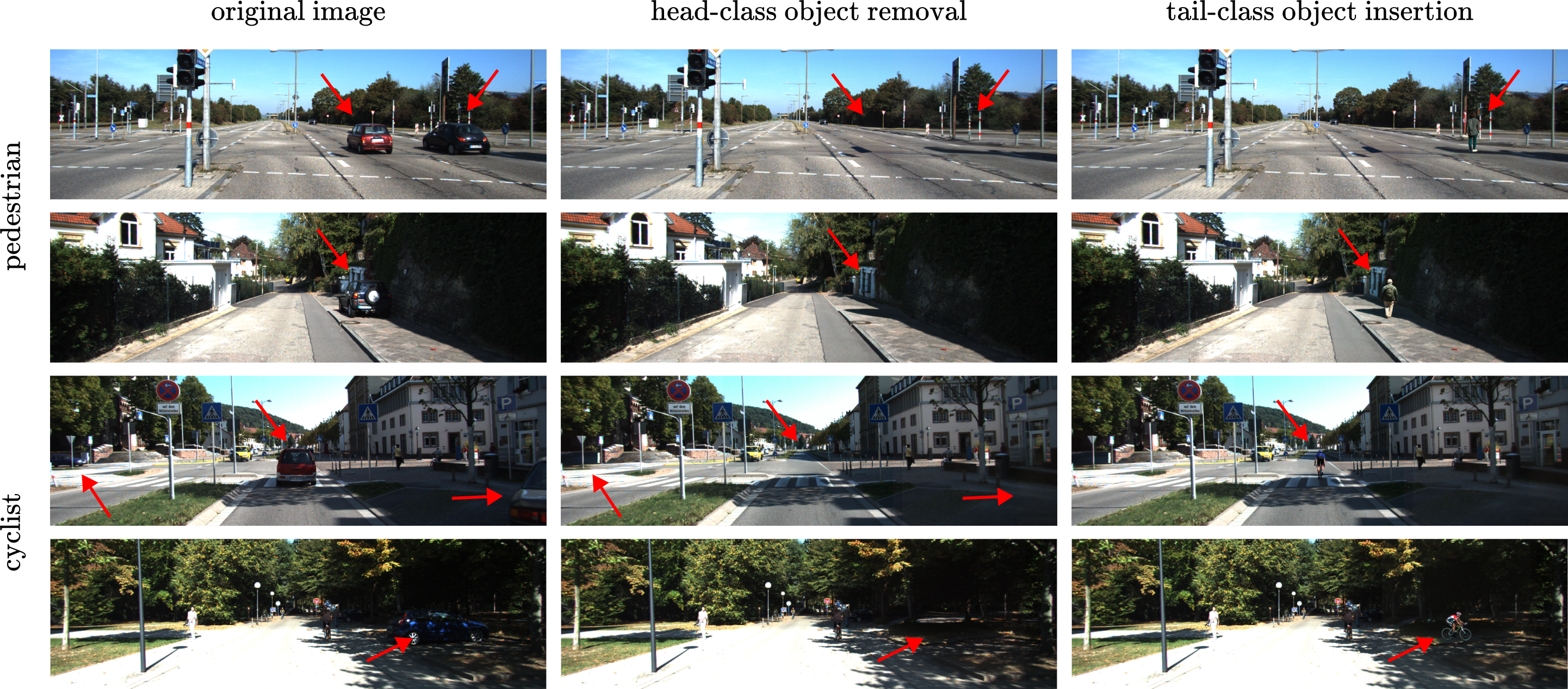}
    \caption{Representative results from the LTDA-Drive augmentation process. The first column shows the original image before augmentation. The second column presents images with head-class objects removed. The third column displays the final augmented images, where tail-class objects are inserted. The first two rows show the augmentation results for pedestrians, while the last two rows show the results for cyclists. Red arrows highlight the original objects to be removed, the regions where removal occurs, and the inserted objects.}
    \label{fig:qua_results}
    \vspace{-4mm}
\end{figure*}

\section{Experiments and Results}
\label{sec:experiments}

\subsection{Dataset and Experiment Set up}

\subsubsection{KITTI Dataset} 
The KITTI dataset \cite{geiger2013vision} is a benchmark for 3D object detection in autonomous driving, featuring annotated objects captured from real-world driving scenes. Following the Monoflex protocol \cite{zhang2021objects}, we split the KITTI dataset into a training set of 3,712 images and a validation set of 3,769 images, focusing on three object categories: car, pedestrian, and cyclist. The training set contains a total of 17,298 annotated objects from the three classes, where there are 14,357 car, 734 cyclist and 2,207 pedestrian instances. Due to the significantly higher prevalance of the car class compared to the other two, it is considered as the head-class, while cyclist and pedestrian are treated as tail-class objects. A detailed breakdown of object counts is provided in Table~\ref{tab:datacount}.

\subsubsection{Augmented Dataset}
\label{sec:augmented_dataset}
To address the long-tail class imbalance in the KITTI dataset, we utilized the proposed LTDA-Drive method to augment the training images. This process involved removing head-class objects (cars) from the images and replacing them with tail-class objects (pedestrians and cyclists) in the cleared regions. The removable head-class objects were selected based on criteria such as not intersecting with other objects and satisfying minimum size requirements, while the inserted objects were further constrained to remain fully within the image boundaries. The augmentation was carried on to generate 612 new images featuring inserted cyclists and 147 new images featuring inserted pedestrians, with each augmented image receiving 1-4 new instances. This resulted in a dataset of 19,965 objects, improving the class distribution for cyclists from 4.24\% to 7.56\%, while maintaining the original pedestrian ratio. Detailed statistics for original and augmented dataset can be found in Tab.~\ref{tab:datacount}.

\subsubsection{Object Detector and Evaluation Metrics}
To further validate the effectiveness of our augmented dataset, we conducted 3D object detection experiments on the original KITTI dataset and our augmented dataset using the MonoFlex model \cite{zhang2021objects}. We utilized the official MonoFlex implementation and assessed its performance on the same validation set, comparing results from models trained on both the original KITTI dataset and the augmented version. We report performance using four standard metrics: average precision for 3D bounding boxes (AP\textsubscript{3D}), 2D bounding boxes (AP\textsubscript{2D}), bird’s-eye view projections (AP\textsubscript{BEV}), and average orientation similarity (AOS). Results are presented over easy, moderate, hard cases based based on the size of the 2D bounding box of the objects, occlusion and truncation levels. Following~\cite{lin2025drivegen}, the intersection-over-union (IoU) thresholds are set to 0.7, 0.5, and 0.5 for cars, and 0.5, 0.25, and 0.25 for pedestrians and cyclists, respectively.
\subsubsection{Implementation Details}
We employed a text-conditioned diffusion inpainting model based on SDXL \cite{podell2023sdxl} for both head-class object removal and tail-class object insertion. The number of inference steps were configured to 30, with the classifier-free guidance scale $w$ set to 8. For object removal, $m=10$ candidate images were generated, and the final selection consisted of $k=1$ candidate. For object insertion, $m=30$ candidate images were generated, and the final output consisted of $k=1$ image. The backbone of the LLM agent for all tasks was ChatGPT-4.1. LTDA-Drive augmentation tasks were executed on A100 and H200 GPUs, while the downstream MonoFlex object detector was trained on V100 GPUs.

\subsection{Main Results}

\begin{figure*}[htbp]
    \centering
    \includegraphics[width=0.99\textwidth]{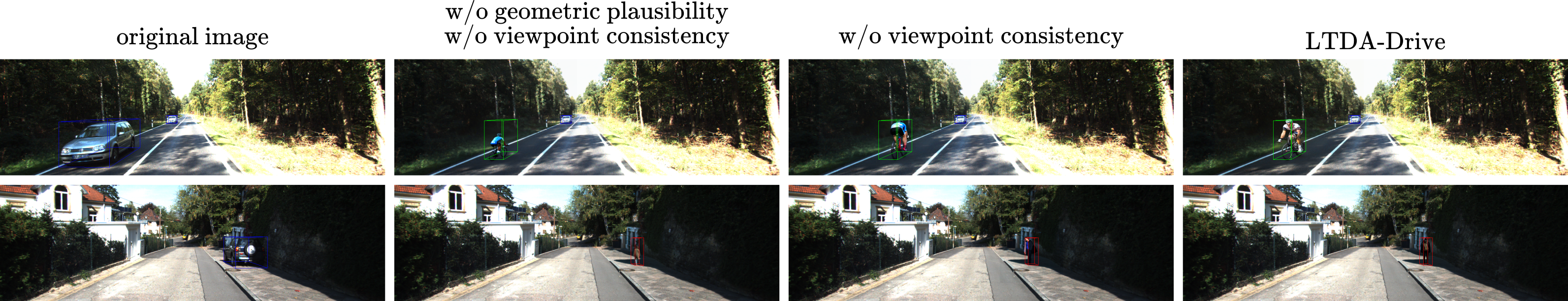}
    \caption{Representative results from the ablation study. The first column displays the original images, while the second and third columns show the ablated versions of LTDA-Drive. The final column illustrates the full LTDA-Drive version. Bounding boxes to be used as ground-truth in downstream 3D object detection are also shown.}
    \label{fig:ablation_results}
    \vspace{-3mm}
\end{figure*}

\subsubsection{Qualitative Results}
Fig.~\ref{fig:qua_results} demonstrates the images generated by the proposed LTDA-Drive framework. As illustrated in the figure, LTDA-Drive successfully removes head-class objects, specifically cars, from the original image, ensuring that background details are retained with minimal disruption. By removing the head-class objects, LTDA-Drive reduces their frequency in the dataset, as demonstrated in Table~\ref{tab:datacount}. Meanwhile, tail-class objects, such as pedestrians and cyclists, are inserted in place of the removed cars to enhance their representation in the dataset. As depicted, both pedestrians and cyclists are positioned in realistic locations, with proper sizes and orientations. As a result, these high-fidelity images form our augmented dataset that can be used to improve the long-tail object detection performance.

\subsubsection{Quantitative Results}
Table~\ref{tab:quant_res} presents the quantitative detection metrics achieved by the MonoFlex models trained on both the original and augmented datasets, with augmentation carried out using the proposed LTDA-Drive method. As indicated, LTDA-Drive consistently outperforms across all metrics and difficulty levels for long-tail classes, including pedestrians and cyclists. Specifically, LTDA-Drive achieves improvements in AP\textsubscript{2D} by 25.1\%, AP\textsubscript{BEV} by 33.2\%, AP\textsubscript{3D} by 34.2\%, and AOS by 46.5\% for cyclists. These results underscore the effectiveness of LTDA-Drive in producing high-quality augmentations that enhance downstream object detection performance.

\subsection{Ablation Study}
To assess the contribution of each component in LTDA-Drive, we conducted ablation studies and presented qualitative results by selectively omitting specific components. Representative images are shown in Fig.~\ref{fig:ablation_results}. Our observations indicate that when the LLM-guided filtering for both geometric plausibility and viewpoint consistency is disabled, the augmented tail-class objects exhibit poor alignment with the bounding boxes and yield mismatches with the orientation of the original objects. Enabling geometric plausibility while disabling viewpoint consistency improves alignment with the bounding boxes, but the orientation of the inserted objects still misaligns with the original object. When both components, geometric plausibility and viewpoint consistency, are included, as in the full LTDA-Drive model, the generated images exhibit high fidelity, with precise alignment to the bounding boxes and the correct orientation.

\section{Conclusion}
% \vspace{-5pt}
\label{sec:conclusion}
In this study, we present LTDA-Drive, a novel data augmentation framework that jointly leverages generative models and LLMs to address the long-tail distribution challenges in perception tasks for autonomous driving. LTDA-Drive performs an in-place replacement augmentation strategy, where frequent head-class objects are removed and replaced with underrepresented tail-class objects via three key modules: 1) head-class object removal, which removes head-class objects from the images while preserving the remaining image content; 2) tail-class object insertion, which seamlessly inserts tail-class objects into the cleared regions, ensuring consistency with the background; and 3) LLM-guided candidate filtering, which assesses the quality of synthesized images. These operations are performed on localized regions, ensuring the preservation of the image’s overall integrity while generating realistic, high-fidelity outputs. Extensive experiments on KITTI dataset demonstrates the efficiency of LTDA-Drive in augmenting long-tail data to improve perception task performance for safety of vulnerable road users.

\textbf{Limitations: }Despite its strong performance, LTDA-Drive has some limitations. First, during the object removal stage, LTDA-Drive is currently restricted to removing only head-class objects that do not intersect with other objects, limiting the number of frequent objects that can be removed. Expanding this capability to remove all head-class objects is an avenue for future work. Second, the current version is designed for augmenting image data. However, applying it to video data—where maintaining consistency across frames is crucial—poses new challenges. Future work will aim to enhance LTDA-Drive to ensure both high accuracy and frame-to-frame consistency in video object detection.

\clearpage
\newpage
\bibliography{custom}

% \appendix

% \section{Example Appendix}
% \label{sec:appendix}

% This is an appendix.

\end{document}